# 深度学习研究综述

张　荣，李伟平，莫　同

北京大学软件与微电子学院，北京　100871

基金项目：国家重点研发计划项目(2017YFC0803609，2017YFB1400400)；河南省交通运输厅科技项目(2016G5)
通信作者：李伟平，wpli@ss.pku.edu.cn

**摘要**

近年来，中美等国家、谷歌等高科技公司纷纷加大对人工智能的投入，深度学习是目前人工智能的重点研究领域之一，本文对深度学习最新进展及未来研究方向进行了分析和总结．首先概述了三类深度学习基本模型，包括多层感知器、卷积神经网络和循环神经网络．在此基础上，进一步分析了不断涌现出来的新型卷积神经网络和循环神经网络．然后本文总结了深度学习在人工智能众多领域中的应用，包括语音处理、计算机视觉和自然语言处理等．最后探讨了深度学习目前存在的问题并给出了相应的可能解决方法．

# Review of Deep Learning

ZHANG Rong, LI Weiping, MO Tong

*School of Software and Microelectronics*, *Peking University*, *Beijing* 100871, *China*

**Abstract**

In recent years, several countries, such as China and the United States, and high-tech companies, such as Google, have increased investment in artificial intelligence. Deep learning is one of the current artificial intelligence research key areas. We analyze and summarize the latest progress and future research directions of deep learning. First, we outline three basic models of deep learning, which are multilayer perceptrons, convolutional neural networks, and recurrent neural networks. On this basis, we further analyze the emerging new models of convolution neural networks and recurrent neural networks. Furthermore, we summarize the applications of deep learning in many areas of artificial intelligence, including speech processing, computer vision, and natural language processing. Finally, we discuss the existing problems of deep learning and provide the corresponding possible solutions.

## 0　引言

2016 年 3 月，“人工智能”一词被写入中国“十三五”规划纲要，2016 年 10 月美国政府发布《美国国家人工智能研究与发展战略规划》文件．Google、Microsoft、Facebook、百度、腾讯、阿里巴巴等各大互联网公司也纷纷加大对人工智能的投入．各类人工智能创业公司层出不穷，各种人工智能应用逐渐改变人类的生活．深度学习是目前人工智能的重点研究领域之一，应用于人工智能的众多领域，包括语音处理、计算机视觉、自然语言处理等．

1943 年，McCulloch 和 Pitts[1] 提出 MP 神经元数学模型．1958 年，第一代神经网络单层感知器由 Rosenblatt[2] 提出，第一代神经网络能够区分三角形、正方形等基本形状，让人类觉得有可能发明出真正能感知、学习、记忆的智能机器．但是第一代神经网络基本原理的限制打破了人类的梦想，1969 年，Minsky[3] 发表感知器专著：单层感知器无法解决异或 XOR 问题；神经网络的特征层是固定的，是由人类设计的，此与真正智能机器的定义不相符．1986 年，Hinton 等[4] 提出第二代神经网络，将原始单一固定的特征层替换成多个隐藏层，激活函数采用 Sigmoid 函数，利用误差的反向传播算法来训练模型，能有效解决非线性分类问题．1989 年，Cybenko 和 Hornik 等[5-6] 证明了万能逼近定理(universal approximation theorem)：任何函数都可以被三层神经网络以任意精度逼近．同年，LeCun 等[7-8] 发明了卷积神经网络用来识别手写体，当时需要 3 天来训练模型．1991 年，反向传播算法被指出存在梯度消失问题．此

后十多年，各种浅层机器学习模型相继被提出，包括 1995 年 Cortes 与 Vapnik[9] 发明的支持向量机，神经网络的研究被搁置. 2006 年，Hinton 等探讨大脑中的图模型[10]，提出自编码器(autoencoder)来降低数据的维度[11]，并提出用预训练的方式快速训练深度信念网[12]，来抑制梯度消失问题. Bengio 等[13] 证明预训练的方法还适用于自编码器等无监督学习，Poultney 等[14] 用基于能量的模型来有效学习稀疏表示. 这些论文奠定了深度学习的基础，从此深度学习进入快速发展期. 2010 年，美国国防部 DARPA 计划首次资助深度学习项目. 2011 年，Glorot 等[15] 提出 ReLU 激活函数，能有效抑制梯度消失问题. 深度学习在语音识别上最先取得重大突破，微软和谷歌[16-17] 先后采用深度学习将语音识别错误率降低至 20% ~30%，是该领域 10 年来最大突破. 2012 年，Hinton 和他的学生将 ImageNet[18] 图片分类问题的 Top5 错误率由 26% 降低至 15%[19]，从此深度学习进入爆发期. Dauphin 等[20] 在 2014 年，Choromanska 等[21] 在 2015 年分别证明局部极小值问题通常来说不是严重的问题，消除了笼罩在神经网络上的局部极值阴霾. 深度学习发展历史如图 1 所示. 图 1 中的空心圆圈表示深度学习热度上升与下降的关键转折点，实心圈圈的大小表示深度学习在这一年的突破大小. 斜向上的直线表示深度学习热度正处于上升期，斜向下的直线表示深度学习热度处于下降期.

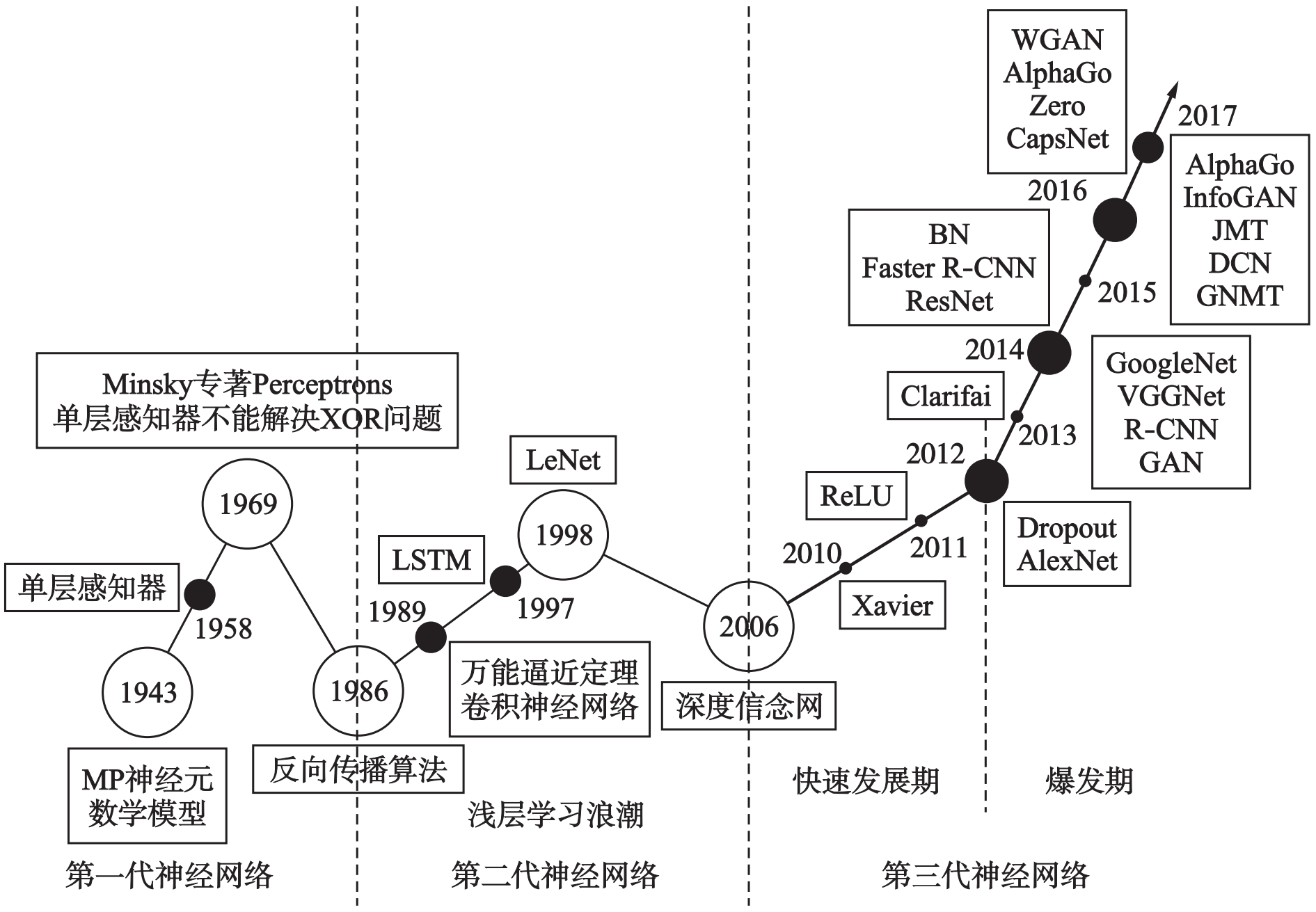

图 1 深度学习发展历史

Fig.1 The history of deep learning

深度学习其实是机器学习的一部分，机器学习经历了从浅层机器学习到深度学习两次浪潮[22]. 深度学习模型与浅层机器学习模型之间存在重要区别. 浅层机器学习模型不使用分布式表示(distributed representations)[23]，而且需要人为提取特征，模型本身只是根据特征进行分类或预测，人为提取的特征好坏很大程度上决定了整个系统的好坏. 特征提取需要专业的领域知识，而且特征提取、特征工程需要花费大量时间. 深度学习是一种表示学习[24]，能够学到数据更高层次的抽象表示，能够自动从数据中提取特征[25-26]. 而且深度学习里的隐藏层相当于是输入特征的线性组合，隐藏层与输入层之间的权重相当于输入特征在线性组合中的权重[27]. 另外，深度学习的模型能力会随着深度的增加而呈指数增长[28].

# 1 基本网络结构

## 1.1 多层感知器

多层感知器(multilayer perception, MLP)[2] 也叫前向传播网络、深度前馈网络，是最基本的深度学习网络结构. MLP 由若干层组成，每一层包含若干个神经元. 激活函数采用径向基函数的多层感知器被称为径向基网络(radial basis function network). 多层感知器的前向传播如图 2 所示.

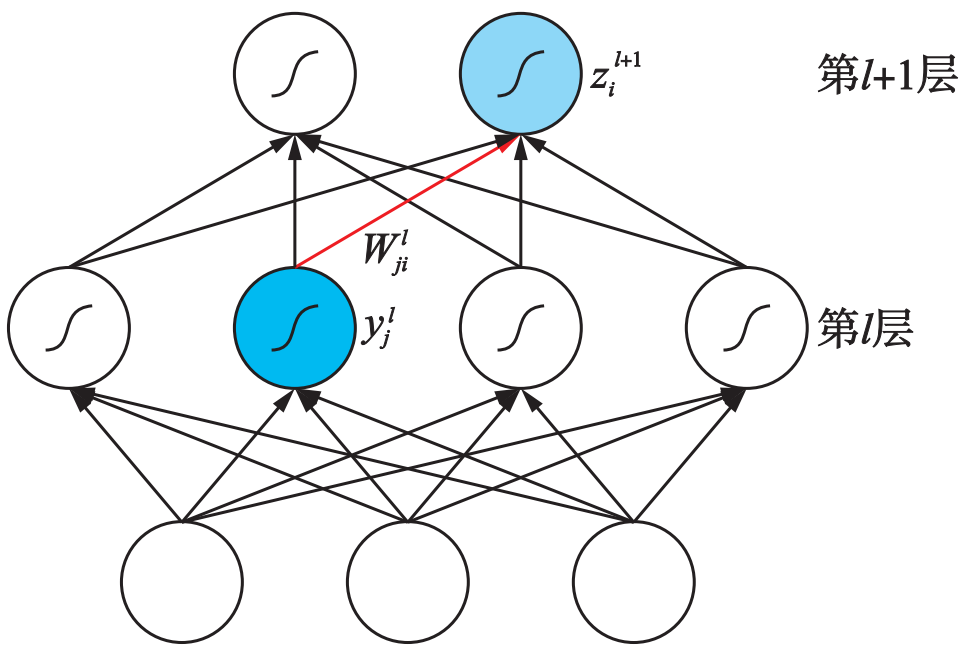

图 2 多层感知器的前向传播

Fig.2 The forward propagation of MLP

MLP 的前向传播公式如式(1)、式(2)所示：

$$z_i^{l+1} = \sum_j W_{ji}^l y_j^l + b_i^l \tag{1}$$

$$y_i^{l+1} = f(z_i^{l+1}) \tag{2}$$

其中，$y_j^l$ 是第 $l$ 层的第 $j$ 个神经元的输出，$z_i^{l+1}$ 是第 $l+1$ 层的第 $i$ 个神经元被激活函数作用之前的值，$W_{ji}^l$ 是第 $l$ 层的第 $j$ 个神经元与第 $l+1$ 层的第 $i$ 个神经元之间的权重，$b_i^l$ 是偏置，$f(\cdot)$是非线性激活函数，常见的有径向基函数、ReLU、PReLU、Tanh、Sigmoid 等.

如果采用均方误差(mean squared error)，则损失函数为

$$J = \frac{1}{2}\sum_i (y_i^L - y_i)^2 \tag{3}$$

其中，$y_i^L$ 是神经网络最后一层第 $i$ 个神经元的输出，$y_i$ 是第 $i$ 个神经元的真实值. 神经网络训练的目标是最小化损失函数，优化方法通常采用批梯度下降法.

### 1.2 卷积神经网络

卷积神经网络(convolutional neural network，CNN)[29]适合处理空间数据，在计算机视觉领域应用广泛. 一维卷积神经网络也被称为时间延迟神经网络(time delay neural network)，可以用来处理一维数据. CNN 的设计思想受到了视觉神经科学的启发，主要由卷积层(convolutional layer)和池化层(pooling layer)组成. 卷积层能够保持图像的空间连续性，能将图像的局部特征提取出来. 池化层可以采用最大池化(max-pooling)或平均池化(mean-pooling)，池化层能降低中间隐藏层的维度，减少接下来各层的运算量，并提供了旋转不变性. 卷积与池化操作示意图如图 3 所示，图中采用 3×3 的卷积核和 2×2 的 pooling.

最早期的卷积神经网络模型是 LeCun 等[29]在 1998 年提出的 LeNet-5，其结构图如图 4 所示. 输入的 MNIST 图片大小为 32×32，经过卷积操作，卷积核大小为 5×5，得到 28×28 的图片，经过池化操作，得到 14×14 的图片，然后再卷积再池化，最后得到 5×5 的图片. 接着依次有 120、84、10 个神经元的全连接层，最后经过 Softmax 函数作用，得到数字 0~9 的概率，取概率最大的作为神经网络的预测结果. 随着卷积和池化操作，网络越高层，图片大小越小，但图片数量越多.

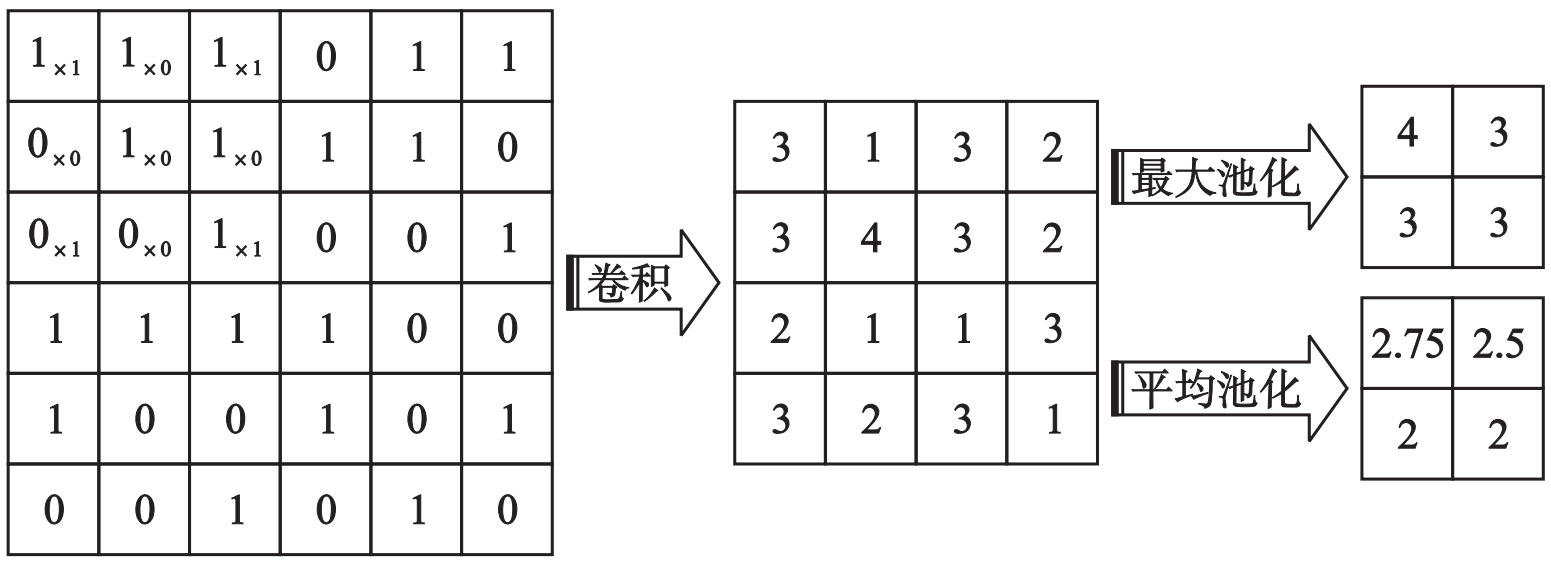

图 3 卷积与池化操作示意图

Fig.3 The illustration for convolution and pooling

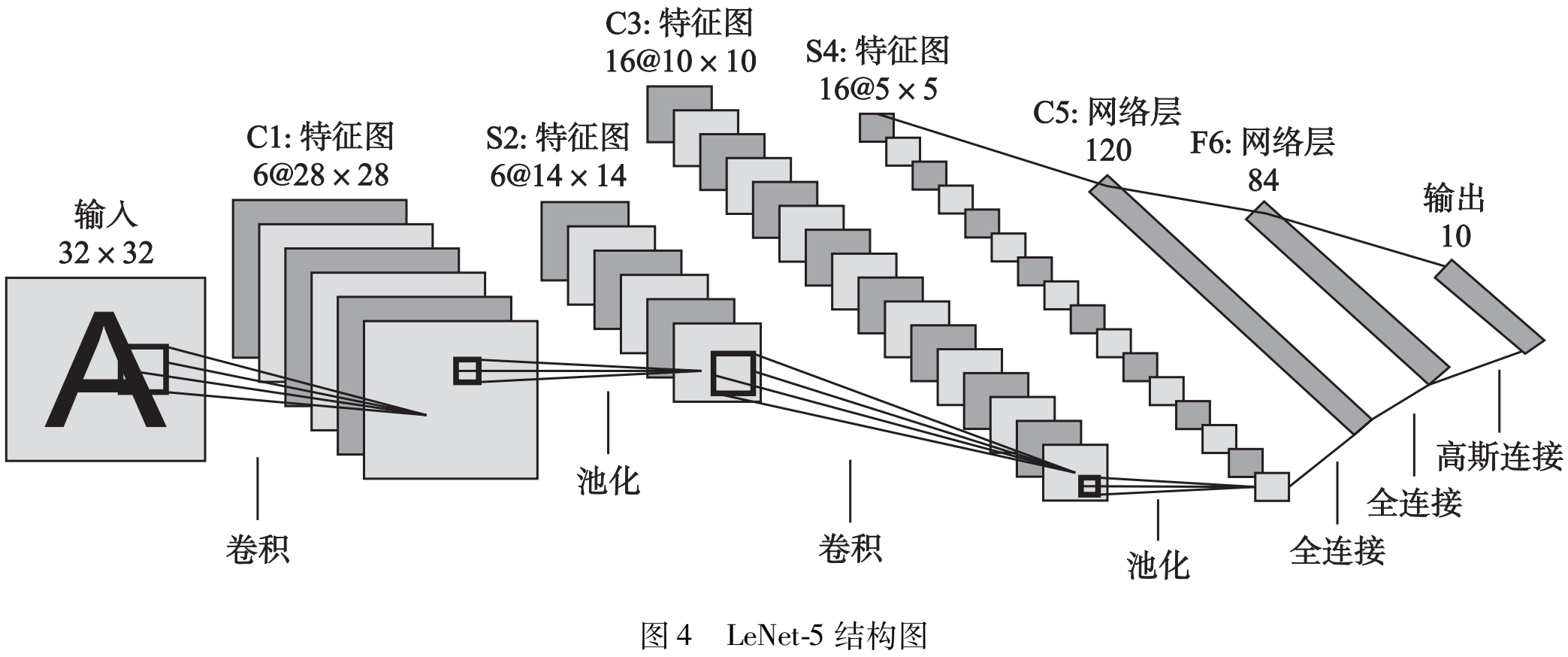

图 4 LeNet-5 结构图

Fig.4 The structure of LeNet-5

CNN 提供了视觉数据的分层表示，CNN 每层的权重实际上学到了图像的某些成分，越高层，成分越具体. CNN 将原始信号经过逐层的处理，依次识别出部分到整体. 可以对 CNN 进行可视化来理解 CNN[30]：CNN 的第二层能识别出拐角、边和颜色；第三层能识别出纹理、文字等更复杂的不变性；第四层能识别出狗的脸、鸟的腿等具体部位；第五层能识别出键盘、狗等具体物体. 比如说人脸识别，CNN 先是识别出点、边、颜色、拐角，再是眼角、嘴唇、鼻子，再是整张脸. CNN 容易在 FPGA 等硬件上实现并获得加速[31]；CNN 同一卷积层内权值共享，都为卷积核的权重. CNN 的局部连接、权值共享、池化操作等特性减少了模型参数[32]，降低了网络复杂性，也提供了平移、扭曲、旋转、缩放不变性.

### 1.3 循环神经网络

循环神经网络(recurrent neural networks，RNN)[4]适合

处理时序数据，在语音处理、自然语言处理领域应用广泛，人类的语音和语言天生具有时序性. RNN 及其展开图如图 5 所示.

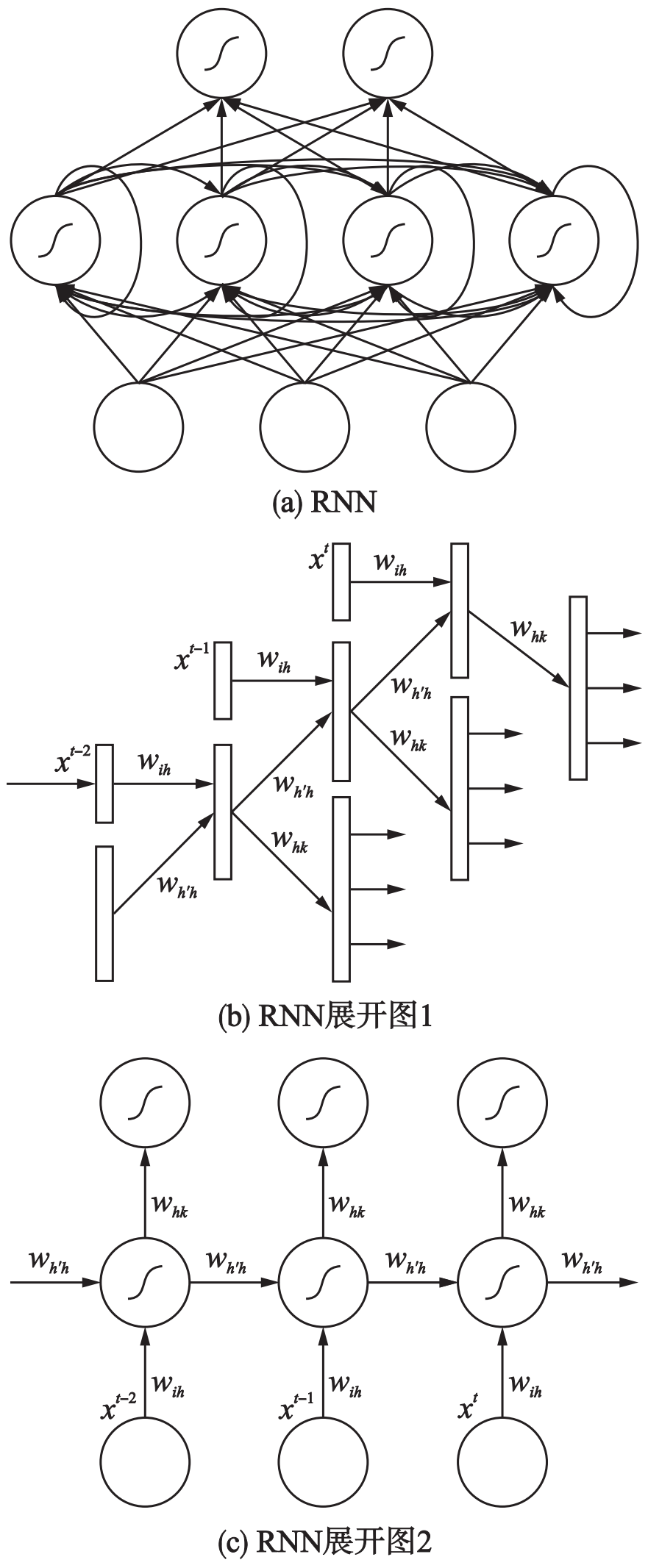

(a) RNN

(b) RNN展开图1

(c) RNN展开图2

图 5　RNN 及其展开图

Fig.5　RNN and its unfolding

RNN 的前向传播公式如式(4) ~ (6)所示：

$$z_h^t = \sum_{i=1}^{I} w_{ih} x_i^t + \sum_{h'=1}^{H} w_{h'h} a_{h'}^{t-1} \tag{4}$$

$$a_h^t = f_h(z_h^t) \tag{5}$$

$$y_k^t = \sum_{h=1}^{H} w_{hk} a_h^t \tag{6}$$

其中，$x_i^t$ 是 $t$ 时刻输入层的第 $i$ 个神经元，$a_{h'}^{t-1}$ 是 $t-1$ 时刻隐藏层的第 $h'$ 个神经元，$z_h^t$ 是 $t$ 时刻隐藏层第 $h$ 个神经元被激活函数作用之前的值，$y_k^t$ 是 $t$ 时刻输出层的第 $k$ 个神经元，$w_{ih}$、$w_{h'h}$、$w_{hk}$ 分别是输入层与隐藏层、隐藏层与隐藏层、隐藏层与输出层之间的权重，$f_h(\cdot)$ 是非线性激活函数.

RNN 将上一时刻隐藏层的输出也作为这一时刻隐藏层的输入，能够利用过去时刻的信息，即 RNN 具有记忆性. RNN 在各个时间上共享权重，大幅减少了模型参数. 但 RNN 训练难度依然较大，因此 Sutskever 等[33] 和 Pascanu 等[34] 都对 RNN 的训练方法进行了改进.

## 2　网络结构改进

### 2.1　卷积神经网络改进

ImageNet[18] 比赛(ImageNet large scale visual recognition competition，ILSVRC)极大促进了卷积神经网络的发展，不断有新发明的卷积神经网络刷新了 ImageNet 成绩. 从 2012 年的 AlexNet[19]，到 2013 年的 ZF Net[30]，2014 年的 VGGNet[35]、GoogLeNet[36]，再到 2015 年的 ResNet[37]，网络层数不断增加，模型能力也不断增强. AlexNet 第一次展现了深度学习的强大能力，ZF Net 是可视化理解卷积神经网络的结果，VGGNet 表明网络深度能显著提高深度学习的效果，GoogLeNet 第一次打破了卷积层池化层堆叠的模式，ResNet 首次成功训练了深度达到 152 层的神经网络. CNN 应用于物体检测的主流方法是 R-CNN[38] 及其之后的改进 Fast R-CNN[39]、Faster R-CNN[40]、Mask R-CNN[41]. 改进的过程其实是用深度学习模型来替代浅层机器学习模型的过程，实现端到端的训练，速度也越来越快. 另外，网中网结构[42] 创新性地提出在网络里嵌套网络的想法；空间变换网络[43] 说明模型效果的提升不一定需要改变网络结构，还可以通过对输入数据进行变换.

1）AlexNet. Hinton 为了验证深度学习的有效性，2012 年参加 ILSVRC 并取得第一名，所用到的神经网络模型被称为 AlexNet. AlexNet 网络包含 5 层卷积层、max-pooling 层和 dropout 层，接着连接 3 层全连接层，最后输出层有 1 000 个神经元，对应 1 000 个分类，经过 Softmax 函数作用后得到每一类的概率. AlexNet 采用平移、翻转、截取图片一部分等方式来增加训练数据，用 dropout 来防止过拟合，用带有动量和权重衰减的批梯度下降方法来训练模型. AlexNet 用两块 GPU 并行训练了 6 天，而且采用 ReLU 作为激活函数比用 Tanh 训练时间缩短了 6 倍. AlexNet 所采用的这一系列技术现在仍然被广泛使用.

2）ZF Net. ZF Net 是 ILSVRC 2013 冠军，错误率为 11.2%，ZF Net 可以认为是 AlexNet 的微调，网络层数仍为 8. Zeiler 和 Fergus 利用反卷积网络对 CNN 进行可视化来理解 CNN 每一层的作用，可视化帮助找到了比 AlexNet 效果更好的网络结构 ZF Net. ZF Net 所需的训练数据更少，AlexNet 用 1 500 万张图片来训练模型，而 ZF Net 只用了 130 万张图片. AlexNet 第一层卷积核为 11 × 11，而 ZF Net 为 7 × 7，卷积核变小使得 ZF Net 在第一层能保留更多的相关信息.

3）VGGNet. Simonyan 等逐次在 AlexNet 中增加卷积层，比较 6 种不同深度的网络，研究网络深度的影响. 结果表明神经网络越深，效果越好，当增加到 16、19 层时，效果提升明显，19 层的网络被称为 VGG-19. VGGNet 严格采用 3 × 3 的卷积核，步长(stride)和填补(padding)都为 1；采用 2 × 2 的 max-pooling，步长为 2. 相比于 ZF Net 7 × 7 的卷积核，VGGNet 卷积核大小只有 3 × 3，使得模型参数更

少，而且连续两层的卷积层使其有 7 ×7 卷积核的效果，之后人们通常也使用 3 ×3 的卷积核. VGGNet 模型用 Caffe 来实现，利用图片抖动来增加训练数据，在图片分类和物体定位任务方面都有很好的效果.

4）GoogLeNet. GoogLeNet 是 ILSVRC 2014 冠军，top 5 错误率为 6.7%，其网络层数为 22 层. GoogLeNet 表明 CNN 不一定是要将卷积层、池化层依次堆叠起来. GoogLeNet 采用 Inception 模块，模块里的卷积层、池化层是并行的，所以不用选择这一层是用卷积层还是池化层. 在 Inception 模块的最后不直接将所有神经元“拉直”排成一排，而是采用池化将 7 ×7 ×1 024 变成 1 ×1 ×1 024，参数量减少到 1/49，GoogLeNet 总的参数量只有 AlexNet 的 1/12. 使用训练好的模型对图片进行分类时，对同一张图片的多张变形图片输出 Softmax 概率后求平均作为此图片的概率.

5）深度残差网络（ResNet）. ResNet 是 ILSVRC 2015 冠军，同一网络赢得图片分类、物体定位、物体检测三项任务冠军，图像分类任务错误率为 3.57%，超过人类错误率 5.1%. ResNet 网络层数达到 152 层，甚至 1 000 层. 深层网络有梯度消失的问题，ResNet 在两层或多层之间直接加上线性连通通路，即构成了残差模块，保证梯度能通过线性通路传到底层，也使得输入层的信息能直接保留到后面网络层.

6）R-CNN. Girshick 等提出 R-CNN 用于完成计算机视觉中的物体检测任务. 物体检测目标是将图片中所有物体用方框框出来，此任务可以分成两个子任务，首先是生成方框将物体框出来，然后对框出来的物体进行分类判断是具体哪个物体. R-CNN 采用选择性搜索（selective search）方法生成大约 2 000 个方框，用已训练好的 CNN 比如 AlexNet 对每一个方框内的图片提取出特征，再将特征放进 SVM 进行分类，同时将特征放入回归器中得到更精确的候选方框.

7）Fast R-CNN. Fast R-CNN 将 R-CNN 中 CNN 提取特征、SVM 分类、回归这三个过程放在一起，形成端到端整体的模型，速度和准确率都得到提升. Fast R-CNN 的输入数据是整张图片和若干方框. 首先用若干卷积层、池化层处理整张图片得到特征图（feature map）；用兴趣区域池化层（region of interest pooling layer）处理每个方框得到固定大小的特征图. 然后接若干全连接层，最后同时输出是某个类别的概率、确定每个类的方框的 4 个值.

8）Faster R-CNN. Faster R-CNN 首先用卷积层、池化层处理整张图片得到特征图，在此特征图上用 region proposal network 来生成方框，其它操作跟 Fast R-CNN 一样. 即 Faster R-CNN 将生成方框的方法也换成了深度学习模型，并由原来在整张图上生成改成在更小的特征图上生成，使得模型训练速度进一步加快.

9）Mask R-CNN. Mask R-CNN 在 Faster R-CNN 基础上增加语义分割的并行分支，在原来生成方框、分类、回归任务基础上增加分割任务，能同时实现物体检测和语义分割. Mask R-CNN 的基础网络使用 ResNeXt-101 和 FPN（feature pyramid network）[44]. 语义分割任务的误差由基于单像素 Softmax 多项式交叉熵变成了基于单像素 Sigmoid 二值交叉熵. Mask R-CNN 加入了 RoIAlign 层，相当于对特征图进行插值.

10）网中网结构（network in network，NIN）. 网中网结构用微型神经网络比如多层感知器，来代替 CNN 中的卷积核，形成了神经网络里嵌套着微型神经网络的结构. 因为已经用微型网络进行了复杂的局部建模，所以 CNN 中最后的全连接层可以由全局 mean-pooling 来代替. 这使得模型参数大大减少，防止了过拟合，也增加了可解释性，NIN 的参数有 2 900 万个，是 AlexNet 的 1/10.

11）空间变换网络（spatial transformer networks，STNs）. 空间变换网络通过变换输入的图片来提升准确率，而不是通过改变网络结构. STNs 里主要包含空间变换模块，其又由本地化网络（localization network）、网格生成器（grid generator）、采样器（sampler）三部分组成. STNs 对于输入的图片，先用本地化网络来预测需要进行的变换，然后网格生成器和采样器对图片实施变换，变换得到的图片被放到 CNN 中进行分类. STNs 的鲁棒性很好，具有平移、伸缩、旋转、扰动、弯曲等空间不变性.

12）其它卷积神经网络改进. 此外，还有其它卷积神经网络改进，包括 deconvolutional networks[45]、stacked convolutional auto-encoders[46]、SRCNN[47]、OverFeat[48]、FlowNet[49]、Mr-CNN[50]、FV-CNN[51]、DeepEdge[52]、DeepContour[53]、deep parsing network[54]、BoxSup[55]、TCNN[56]、3 维 CNN[57] 等.

### 2.2　循环神经网络改进

循环神经网络存在梯度消失或者梯度爆炸问题[58]，无法利用过去长时间的信息，例如当激活函数是 Sigmoid 函数时，其导数是个小于 1 的数，多个小于 1 的导数连乘就会导致梯度消失问题. LSTM[59]、分层 RNN[60-61] 都是针对这个问题的解决方案. RNN 一般只能处理时间序列等一维数据，多维 RNN[62] 被提出来处理图像等多维数据. 针对很多自然语言处理任务都需要利用上下文信息，双向 RNN[63] 通过双向处理同一个序列来利用上下文信息. RNN 存在训练算法复杂、计算量大的问题，回声状态网络[64] 不需要反复计算梯度就能达到很高的精度，为 RNN 的训练提供了新思路. 循环神经网络缺乏推理功能，无法完成需要推理的任务，神经图灵机[65] 与记忆网络[66] 通过增加记忆模块来解决此问题.

1）长短时记忆网络（long short-term memory，LSTM）. 长短时记忆网络用 LSTM 单元替代 RNN 中的神经元，在输入、输出、忘记过去信息上分别加了输入门、输出门、遗忘门来控制允许多少信息通过. LSTM 有单元状态（cell state）和隐藏状态（hidden state）两个传输状态，单元状态随时刻改变缓慢，而不同时刻的隐藏状态可能会很不同. LSTM 建立了门机制来达到旧时刻输入与新时刻输入之间的权衡，本质是根据训练目标，调整出记忆的重点然后进行整串编码. LSTM 能够记住需要长时间记忆的，忘记不重要的，能缓解梯度消失、梯度爆炸问题，在较长的序列上比 RNN 有更好的表现.

2）GRU（gated recurrent unit）. GRU[67]是 LSTM 的轻量级变体，只有两个门：更新门和重置门. 更新门决定保留过去多少信息，以及从输入层输入多少信息；重置门与 LSTM 里的遗忘门类似. GRU 没有输出门，所以总是输出完整状态. GRU 在所有的地方使用了更少的连接，参数更少，所以训练速度更容易更快.

3）分层 RNN（hierarchical RNN）. 分层 RNN 利用了时间依赖是分层结构化的先验知识，长时间依赖能够由一个变量来表示. 分层多尺度 RNN（hierarchical multiscale RNN）[61]通过编码不同时间尺度的时间依赖来捕捉序列中潜在的分层结构. 分层 RNN、分层多尺度 RNN 都能缓解梯度消失问题.

4）双向 RNN（bidirectional RNN）. 双向 RNN 向前、向后双向处理同一个序列，相当于有两个 RNN，然后接着同一个输出层. 向前和向后隐藏层之间没有连接，避免形成信息回路. 比如对于自然语言处理任务，单词的上文跟下文都会对单词有影响，双向 RNN 就能利用单词的上下文信息. 双向 RNN 能同时捕捉到前后的信息，双向 LSTM[68-69]、双向 GRU[70]也被证明非常有效.

5）多维 RNN（multi-dimensional RNN）. RNN 适合处理时间序列等一维数据，多维 RNN 能够处理多维数据，包括图像、视频、医学成像等. 多维 RNN 的思想是将多维数据按照一定顺序排列成一维数据，进而能用 RNN 处理，当然这种排列顺序需要保持多维数据的空间连续性.

6）回声状态网络（echo state network，ESN）. ESN 将 RNN 中的隐藏层变成了存储池，存储池内的神经元之间随机连接，存储池即隐藏层不是整齐划一的神经元网络层. 输入会回荡在存储池中，就像有回声一样，故此网络被称为回声状态网络. ESN 输入层到隐藏层、隐藏层到隐藏层的权重随机初始化后固定不变. 模型训练的时候，只更新隐藏层到输出层的权重，即变成了线性回归问题，所以训练速度快.

7）神经图灵机（neural Turing machines）. 神经图灵机本质是使用外部存储矩阵来进行交互的 RNN，整体系统与图灵机类似，用神经网络的方法建立了"输入纸带到输出纸带的映射". 控制器网络（controller network）通过并行的读写头改变外部存储矩阵的值来读取或写入记忆. 控制器网络会接收 RNN 的输入，也会输出给 RNN.

8）记忆网络（memory networks）. 记忆网络主要包含记忆单元和输入、生成、输出、应答四个模块. 记忆单元其实是一个数组，数组的每一个元素保存一句话的记忆. 记忆网络可以回答需要复杂推理的问题[71]，实现自动问答. 输入的文本被输入模块编码成特征向量，生成模块根据此向量对记忆单元进行读写操作，对记忆进行更新. 输出模块会将记忆按照与问题的相关程度加权组合得到输出向量，最终应答模块根据输出向量编码生成问题的答案. 神经图灵机与记忆网络都适合完成需要推理与符号处理的任务.

9）其它循环神经网络改进. 其它循环神经网络改进也主要是两个方向，一个是改进网络结构，另一个是额外加入记忆模块. 结构方面有：GFRNN[72]、stacked RNN[73]、Tree-Structured LSTM[74]、grid LSTM[75]、segmental RNN[76]、seq2seq for sets[77]等. 记忆方面有：neural GPU[78]、pointer network[79]、deep attention recurrent Q-Network[80]、dynamic memory networks[81]等.

### 2.3 其它改进

有些创新能同时改进卷积神经网络和循环神经网络，比如批标准化与 attention 等. 另外有些有意思的应用同时用到了卷积神经网络与循环神经网络.

1）批标准化（batch normalization）. 训练过程权重更新会改变网络层输入的分布，导致需要更低的学习率与精细的权重初始化，训练时间也会变长. 对每一网络层的输入进行批标准化能保证输入分布的统一，并在一定程度上能替代 dropout. 批标准化首先被应用于卷积神经网络[82]，而后也被应用于循环神经网络[83]中.

2）基于 attention 的模型[84]. 人在看东西时目光会集中于感兴趣的部分，当翻译文章时我们关注当前翻译的那部分，而不是整篇文章，这些都说明人类系统中存在注意力（attention）. 类似地，attention 机制使得深度学习模型能够只集中关注输入数据中最为重要的一部分，相应的模型有基于 attention 的 CNN[85]、基于 attention 的 RNN[86]和基于 attention 的 LSTM[87]. Attention 其实是在网络中加入了关注区域的移动、缩放、旋转机制，此采用强化学习来实现.

3）卷积神经网络与循环神经网络结合. 卷积神经网络与循环神经网络结合可以做出很有意思的应用，比如根据图片生成描述文字[88]等. 根据图片生成描述文字，首先是用卷积神经网络对图片信息进行编码，得到其固定长度的向量表示. 然后用循环神经网络对此向量进行解码，生成图片对应的描述文字.

## 3 深度学习典型应用

### 3.1 语音处理

深度学习最先在语音处理领域取得突破性进展[16]，无论是在标准小数据集上[89]还是大数据集上[17]. 语音处理领域主要有两大任务：语音识别和语音合成. 深度学习广泛应用于语音识别[90]中，Google[91]推出端到端的语音识别系统、百度[92]推出语音识别系统 Deep Speech 2. 2016 年，微软[93]在日常对话数据上的语音识别准确率达到 5.9%，首次达到人类水平. 各大公司也都用深度学习来实现语音合成，包括 Google[94]、Apple[95]、科大讯飞[96]等. Google DeepMind[97]提出并行化 WaveNet 模型来进行语音合成，百度[98]推出产品级别的实时语音合成系统 Deep Voice 3.

### 3.2 计算机视觉

深度学习被广泛应用于计算机视觉各种任务，包括交通标志检测和分类[99]、人脸识别[100]、人脸检测[101]、图像分类[37]、多尺度变换融合图像[102]、物体检测[41]、图像语义分割[103]、实时多人姿态估计[104]、行人检测[105]、场景识别[106]、物体跟踪[56]、端到端的视频分类[107]、视频里的人体动作识别[108]等. 另外，还有一些很有意思的应用，如给黑白照片自动着彩色[109]、将涂鸦变成艺术画[110]、艺术风格转移[111]、去掉图片里的马赛克[112]等. 牛津大学和

Google DeepMind[113]还共同提出了 LipNet 来读唇语，准确率达到 93%，远超人类 52% 的平均水平.

### 3.3 自然语言处理

NEC Labs America[114]最早将深度学习应用于自然语言处理领域. 目前处理自然语言时通常先用 word2vec[115]将单词转化成词向量，其可以作为单词的特征[4]. 自然语言处理领域各种任务广泛用到了深度学习技术，包括词性标注[81]、依存关系语法分析[116]、命名体识别[117]、语义角色标注[118]、只用字母的分布式表示来学习语言模型[119]、用字母级别的输入来预测单词级别的输出[119]、Twitter 情感分析[120]、中文微博情感分析[121]、文章分类[122]、机器翻译[123]、阅读理解[124]、自动问答[71-81]、对话系统[125]等.

### 3.4 其它

在生物信息学方面，深度学习能够被用来预测药物分子的活动[126]，预测人眼停留部位[127]，预测非编码 DNA 基因突变对基因表达与疾病的影响[128]. 金融行业积累了大量的数据，故深度学习在金融方面也有众多应用，包括金融市场预测[129]、证券投资组合[130]、保险流失预测[131]等，也相应涌现出一批金融科技创业公司. 另外，基于深度学习的实时发电调度算法[132]能在满足实时发电任务的前提下，使机组总污染物排放量降低，达到节能减排的目的. 深度学习还可以诊断电动潜油柱塞泵的故障[133]，避免故障事故的发生，有效延长检泵周期. 深度学习应用于强非线性、复杂的化工过程软测量建模[134]中也能获得很好的精度.

## 4 深度学习目前问题及进一步研究方向

虽然深度学习使得诸多领域取得突破性进展，但是深度学习仍然存在一些问题，攻克解决这些问题是学者们的进一步研究方向，本文针对这些问题也探讨了相应的可能解决思路. 问题被分为训练问题、落地问题、功能问题和领域问题，训练问题指的是深度学习训练时间太长等问题，落地问题指的是限制了深度学习实际落地应用的问题，功能问题指的是深度学习目前还不能很好完成的任务，领域问题指的是针对计算机视觉和自然语言处理特定领域的问题.

### 4.1 训练问题

1）训练时间太长，需要大量训练计算资源. 在 WMT'14 英语到法语的数据集上，Google[135]用 96 块 NVIDIA K80 GPU 需要训练6天来得到基本的机器翻译模型，另外还需要 3 天来微调进一步改进模型. 这还只是训练一次模型，再加上需要调各种超参数，总的训练时间非常长，而且 96 块 K80 GPU 成本也非常高. 深度学习模型需要的计算资源太多了，训练时间也太长了，需要全新的硬件、算法、系统设计来加速模型的训练. 例如硬件方面不只采用具有通用性的 GPU，还可以采用高性能低功耗的可编程可配置型 FPGA 芯片、为了某种特定需求而专门定制的 ASIC 芯片. Google 为 TensorFlow 深度学习框架专门设计了 ASIC 芯片 TPU 及二代 Cloud TPU. 自动驾驶领域也倾向于采用 ASIC 芯片，Nvidia 发布针对 L5 级全自动驾驶的 AI 处理器 Drive PX Pegasus.

2）梯度消失问题，训练难度大. 深层模型的训练难度非常大，网络层数太多的模型存在梯度消失问题. 激活函数例如 Sigmoid 函数的导数是个很小的数，多个很小的数连乘之后几乎为 0，则梯度无法从输出层传到输入层. 可以从训练方法、技巧、网络结构等方面来缓解梯度消失问题. 训练方法上：Hinton 等[12]提出先预训练后微调的训练方法，先无监督逐层预训练，再用反向传播算法对整个网络进行微调训练，预训练每次只训练一层隐藏层避免了梯度消失问题. 技巧上：用 ReLU[15]来代替 Sigmoid 作为激活函数、使用 dropout、批标准化（batch normalization）[82]，这些技巧都能缓解梯度消失问题. 网络结构上：LSTM 采用门机制，加了输入门、输出门、遗忘门来控制允许多少信息通过；highway networks[136]加入携带门（carry gate）和转化门（transform gate）使得输出由直接输入和转化后的输入两部分组成；ResNet[37]在在两层或多层之间直接加上线性连通通路，保证梯度能通过线性通路传到底层.

3）依赖大规模带标签训练数据. 深度学习严重依赖大规模带标签数据来训练模型. 人工数据标注耗时耗力，代价高昂；某些特定领域如疑难杂症等，几乎不可能收集到足够多的带标签数据. 因此无监督学习是深度学习接下来的一个重要研究方向，目前已经有一些成果，包括 Goodfellow 等[137]提出的生成对抗网络（generative adversarial nets，GANs）、微软[138]提出的新的学习范式对偶学习.

4）分布式训练问题. 深度学习模型在单机上训练时间过长，尤其是当训练数据太多时，而且规模过大的模型也无法放入一台机器里. 所以需要进行大规模分布式训练深度学习模型. 分布式并行训练可以分为数据并行、模型并行和混合并行. 混合并行里同时使用了数据并行和模型并行，例如可以在不同机器间使用数据并行，同一台机器上使用模型并行. 数据并行是目前多数分布式系统的首选，各种数据并行方法的区别在于是参数平均法还是更新式方法、是同步更新还是异步更新、是中心化同步还是分布式同步. 参数平均法是传输各节点的参数到参数服务器，然后所有参数求平均得到全局参数，更新式方法传输的不是参数，而是参数的更新量. 微软[139]提出延迟补偿的异步更新方法，在同步更新与异步更新中寻找合适的平衡，并开源了参数服务器框架 Multiverso. Strom[140]去掉中心的参数服务器，平等地在各节点间传输参数更新量，高度压缩节点间的更新使得网络通信减少了 3 个数量级.

### 4.2 落地问题

1）对抗性样本攻击. Szegedy 等[141]发现深度学习会被对抗性样本攻击，即本来正确分类的图片加上小的扰动能使深度学习模型误判成别的类别. 知道神经网络结构的攻击被称为白盒攻击（white box attack）；对抗性样本还可以迁移，一个模型产生的对抗性样本也会被另一模型错误分类，此为黑盒攻击（black box attack）. 对抗性样本攻击说明深度学习并没有那么可靠，Goodfellow 等[142]认为这是由于深度学习模型在高维空间中的线性性质导致的. 为了防止对抗性样本攻击，可以进行对抗性训练，将对抗性样

本跟普通样本都作为训练数据来训练模型，使用对抗目标函数，能使模型更加正则化.

2）鲁棒性差. 即使深度学习的平均准确率很高，但在某些测试用例上的预测效果可能很差. 高可靠性系统如无人车，远程外科手术，卫星发射等，不允许出现某个很离谱的结果. 故需要提高深度学习的鲁棒性，保证坏的时候不至于太坏，使得深度学习的应用领域更广泛.

3）太多的超参数. 针对实际问题，如何设计一个最适合的深度模型？深度学习有太多的超参数，包括如何进行数据采集、生成、选择、划分；神经网络结构是用 MLP、CNN 还是 RNN；神经网络层数、每层的神经元数量；权重初始化方法；正则项系数（weight decay）；动量（momentum）、学习率、learning rate decay、dropout；迭代次数、batch；模型更新规则是用 SGD 还是 Adam；分布式训练结果如何聚合等等. 可以用另一个神经网络来学习这些超参数，让神经网络设计变得自动化. Google DeepMind[143] 采用 Learning to learn 算法，用另一个网络来调整学习率，使得收敛更快. Google 发布了能自动搜索最优网络结构的 Cloud AutoML. 韩红桂等[144] 提出利用竞争机制来动态增加或删减隐藏层神经元，动态优化神经网络结构. 张昭昭等[145] 提出多层自适应模块化神经网络结构设计方法.

4）可解释性差. 深度学习模型的可解释性差，是典型的黑箱算法，模型复杂，通常包含上亿个参数. 线上应用模型后，如果对某个用户造成严重影响，无法确定是哪个参数出了问题，从而无法针对性地调整某个参数来解决此用户的问题. 而且模型的可解释性问题限制了其在医学、自动驾驶、军事、航天等重要领域的应用. 可以尝试将深度学习与符号学习、可解释性算法进行结合，使其既有深度学习强大的表达能力，又有一定的可解释性. 微软用图学习机（Graph Engine）来统一机器学习与知识图谱. Stanford 博士生 Wu 等[146] 将深度学习与可解释性决策树算法结合起来，用树正则化方法来提高深度学习的可解释性.

5）模型太大. 深度学习模型本身非常大，不方便放在 GPU 中，更不方便在移动端使用. 特别对于语言模型来说，词表非常大，输出神经元很多，导致模型非常大. 所以需要在保证准确率的前提下，进行模型压缩，将模型变小. 模型压缩方法主要有参数修剪和共享[147]、低秩分解[148]、压缩卷积滤波器[149]、知识精炼[150] 四类[151]. 参数修剪和共享是去除对准确率没有提升的冗余参数，根据减少信息冗余或参数空间冗余的方式，参数修剪和共享又可以细分为量化和二进制、剪枝和共享、设计结构化矩阵三类. 低秩分解是用矩阵分解或张量分解来评估最具信息量的参数；压缩卷积滤波器是设计特殊结构的卷积滤波器来减少存储和计算的复杂度；知识精炼是提取大型模型里的知识来训练更小更紧凑的模型.

### 4.3 功能问题

1）不能像人类一样进行小样本学习. 深度学习需要大规模的训练数据，样本利用率不高，但是人类的学习只需要极少几个样本. 其实小孩在学习过程中，利用了大人传授的知识. 目前还缺乏统一的框架向深度学习模型提供领域先验知识. 要像人类一样进行小样本学习，可以尝试将深度学习与知识图谱、逻辑推理、符号学习等结合在一起，同时利用好数据与知识.

2）不能很好完成动态决策性任务. 决策性任务会随着时间而改变，是动态的；而且往往与环境有复杂的交互，各种因素互相影响. 金融股票预测是个典型的动态决策性任务，同一股票同一价格在不同的时间可能就需要进行相反的买卖操作；一只股票的涨跌会对其他股票产生影响. 动态决策性任务内部之间互相博弈，可以尝试将深度学习与博弈论结合来完成动态决策性任务.

3）不能很好完成逻辑推理任务. 深度学习目前只是在图像识别、语音识别等感知层面的任务有较好的表现，缺乏逻辑推理能力，无法很好地完成需要逻辑推理的任务. 可以考虑将深度学习与擅长逻辑推理的符号学习、存储了知识的知识图谱结合；另外还可以给深度学习模型增加记忆模块，如神经图灵机[65] 与记忆网络[66] 等.

4）不能很好处理已有特征的小数据问题. 深度学习由于能够自动提取特征，所以在图像识别等人为很难提取特征的任务上表现很好. 但是一些任务如保险用户流失预测等，人工能够很好地提取有效特征，而且训练数据较少. 深度学习在这些已有特征的小数据问题上效果还不如 GBDT、XGBoost、LightGBM[152] 等集成学习算法，甚至不如普通的浅层机器学习算法[153].

5）无法同时处理多任务. 人脑是多才多艺的，能同时识别语音，识别图像，理解文字等. 而目前深度学习模型都是针对某一特定任务用特定数据集训练的，训练得到的模型也只能完成这一特定任务. 为了能完成多任务，进一步实现通用人工智能，可以尝试将不同功能的神经网络以某种方式连接成更大的神经网络. Google[154] 通过稀疏门矩阵将多个多层感知器子网络组合成超大网络，用反向传播同时训练所有子网络.

6）终极算法. 机器学习有五大流派：符号主义、连接主义、进化主义、贝叶斯主义、分析主义，深度学习属于连接主义. 深度学习的扩展性很强，可以尝试以深度学习为基础，将其他流派包含进来，形成集成了五大流派的终极算法（master algorithm）[155]. OpenAI[156] 发现用进化主义的遗传算法替代反向传播算法来训练深度强化学习能更快收敛. Google[157] 开源了概率建模推理库 Edward，清华大学[158] 开源了贝叶斯深度学习的概率编程库 ZhuSuan，Uber 和 Stanford 开源了深度概率编程库 Pyro，这些都证明了贝叶斯主义与深度学习可以结合在一起. 另外，深度学习还可以跟逻辑回归等广义线性模型结合[131].

### 4.4 领域问题

1）图像理解问题. 深度学习目前在图像识别等感知任务有较好的表现，但在图像理解如视觉关系理解、图片内容问答、视觉注意点预测等方面成果还并不多. 视觉关系理解首先需要检测出关键对象，然后预测对象之间的关系. 图片内容问答是根据给定的图片，回答相应的问题. 视觉注意点预测是对于给定的图片，预测人最感兴趣图片的哪一部分. 这些都需要对图像内容有很好的理解，图像

理解问题需要学者们的进一步探索.

2）自然语言处理问题. 语言其实是比语音、图像更高级的非自然信号，是完全由人脑产生和处理的符号系统. 深度学习在自然语言处理上的效果还不像语音、图像那么显著，但是深度学习是受人脑启发得到的算法，相信深度学习接下来会在自然语言处理领域有更多的成果.

## 参考文献

（下转第 410 页）

## 作者简介

何文韬(1994－), 男, 硕士生. 研究领域为复杂工业过程建模与集成优化控制理论.
邵　诚(1958－), 男, 博士, 教授, 博士生导师. 研究领域为复杂工业过程建模与集成优化控制理论.

---

(上接第397页)

## 作者简介

张　荣(1992－), 男, 博士生. 研究领域为深度学习.
李伟平(1973－), 男, 博士, 教授, 博士生导师. 研究领域为服务计算, 机器学习.
莫　同(1981－), 男, 博士, 副教授, 硕士生导师. 研究领域为数据挖掘.